%% file: root.tex
\documentclass[letterpaper, 10 pt, conference]{IEEEtran}  % Comment this line out if you need a4paper

\input{setup}

\title{\LARGE \bf
SymAware: A Software Development Framework for Trustworthy Multi-Agent Systems with Situational Awareness$^*$
}

\author{Ernesto Casablanca$^{1}$, Zengjie Zhang$^{2}$, Gregorio Marchesini$^{3}$, \\
Sofie Haesaert$^{2}$, Dimos V. Dimarogonas$^{3}$, and Sadegh Soudjani$^{4}$
\thanks{*This work was supported by the European project SymAware under grant No. 101070802. We thank the whole SymAware consortium for their input during discussions.
The members of the project are available at \url{https://www.symaware.eu/}.}
\thanks{$^{1}$School of Computing, Newcastle University, Newcastle, United Kingdom.  E-mail: {\tt\small e.casablanca2@ncl.ac.uk}.}%
\thanks{$^{2}${Department of Electrical Engineering, Eindhoven University of Technology, Eindhoven, The Netherlands. E-mail: {\tt\small \{z.zhang3, s.haesaert\}@tue.nl}.}}%
\thanks{$^{3}$Division of Decision and Control Systems, KTH Royal Institute of Technology, Stockholm, Sweden. E-mail: {\tt\small \{gremar,dimos\}@kth.se}.}%
\thanks{$^{4}${Max Planck Institute for software systems. E-mail: {\tt\small sadegh@mpi- sws.org}.}}%
}

\begin{document}

\maketitle
\thispagestyle{empty}
\pagestyle{empty}

%%%%%%%%%%%%%%%%%%%%%%%%%%%%%%%%%%%%%%%%%%%%%%%%%%%%%%%%%%%%%%%%%%%%%%%%%%%%%%%%

\begin{abstract}
    Developing trustworthy multi-agent systems for practical applications is challenging due to the complicated communication of situational awareness (SA) among agents. This paper showcases a novel efficient and easy-to-use software framework for multi-agent simulation, named SymAware which provides a rich set of predefined data structures to compute, store, and communicate SA for agents. It also provides an abstract interface for the agents to compute their control inputs taking into account the awareness of the situation, knowledge, and risk of surrounding agents. Besides, utilizing a cluster of specialized components, SymAware hides the heavy computation of physical rendering and communication interfacing of simulation engines behind the control threads, resulting in high implementation efficiency in bridging the gap between conceptual prototyping and practical applications. Three multi-agent case studies are used to validate the efficacy and efficiency of this software framework.
\end{abstract}

%%%%%%%%%%%%%%%%%%%%%%%%%%%%%%%%%%%%%%%%%%%%%%%%%%%%%%%%%%%%%%%%%%%%%%%%%%%%%%%%
\section{Introduction}

Modern cyber-physical systems must reach a higher level of autonomy to meet the increasing demand for safety and efficiency, which proposes higher requirements on seamless coordination and collaboration between the intelligent agents intended for different functionalities~\cite{zhang2021physical, karnouskos2020industrial}.
For example, an autonomous driving system needs to continually monitor the situations of surrounding vehicles and help the human driver perform safe maneuvers~\cite{ning2021survey}.
An intelligent warehouse should maintain a group of robots and resolve their faults and failures  automatically~\cite{tubis2023intelligent}.
An intelligent traffic management system aims at mitigating accident risks via automatic reasoning of human mistakes~\cite{scala2019tackling}.
Most of these applications have been investigated under the paradigm of \gls{mas}, a complex system with multiple dynamic components connected via a certain form of communication~\cite{dorri2018multi}.
Abundant work has been devoted to the design~\cite{paccagnan2022utility}, communication~\cite{berna2004communication}, coordination control~\cite{liu2024distributed}, simulation~\cite{michel2018multi}, and test~\cite{houhamdi2011multi} of \glspl{mas}.
Nevertheless, applying these \glspl{mas} approaches to complicated practical problems is still severely limited due to the lack of trust.

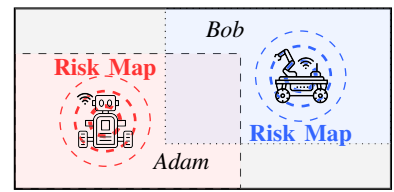
\begin{figure}[htbp]
    \noindent
    \hspace*{\fill}
    \begin{tikzpicture}[scale=1,font=\small]

        \definecolor{darkred}{RGB}{255, 51, 51}
        \definecolor{darkblue}{RGB}{51, 101, 255}
        \definecolor{darkgreen}{RGB}{51, 153, 51}

        \definecolor{shadowyellow}{RGB}{255, 255, 229}
        \definecolor{shadowgreen}{RGB}{217, 242, 217}
        \definecolor{shadowred}{RGB}{255, 204, 204}
        \definecolor{shadowblue}{RGB}{204, 221, 255}

        \definecolor{shadowgray}{RGB}{217, 217, 217}

        %%%%%%%%%%%%%%%%%%%%%%%%%%%%%%%%%%%%%%%%%%%%%%%%
        \node[minimum height=2.4cm, minimum width=5cm, draw,thick] (warehouse) at (0cm,-0.3cm) {};

        \node[minimum height=1.8cm, minimum width=3cm, draw, dashed, fill=shadowred, fill opacity=0.3] (Arange) at (-1cm,-0.6cm) {};
        \node[minimum height=0.6cm, minimum width=2cm, fill=shadowgray, fill opacity=0.3] () at (-1.5cm,0.6cm) {};

        \node[minimum height=1.8cm, minimum width=3cm, draw, dotted, fill=shadowblue, fill opacity=0.3] (Brange) at (1cm,0cm) {};
        \node[minimum height=0.6cm, minimum width=2cm, fill=shadowgray, fill opacity=0.3] () at (1.5cm,-1.2cm) {};

        \draw[dashed, color=darkred, line width=1.5pt] ([xshift=-0.3cm] Arange.center) circle (0.2cm);
        \draw[dashed, color=darkred, line width=1pt] ([xshift=-0.3cm] Arange.center) circle (0.4cm);
        \draw[dashed, color=darkred, line width=0.5pt] ([xshift=-0.3cm] Arange.center) circle (0.6cm);
        \node[] (Adam) at ([xshift=-0.3cm] Arange.center) {\includegraphics[width=0.8cm]{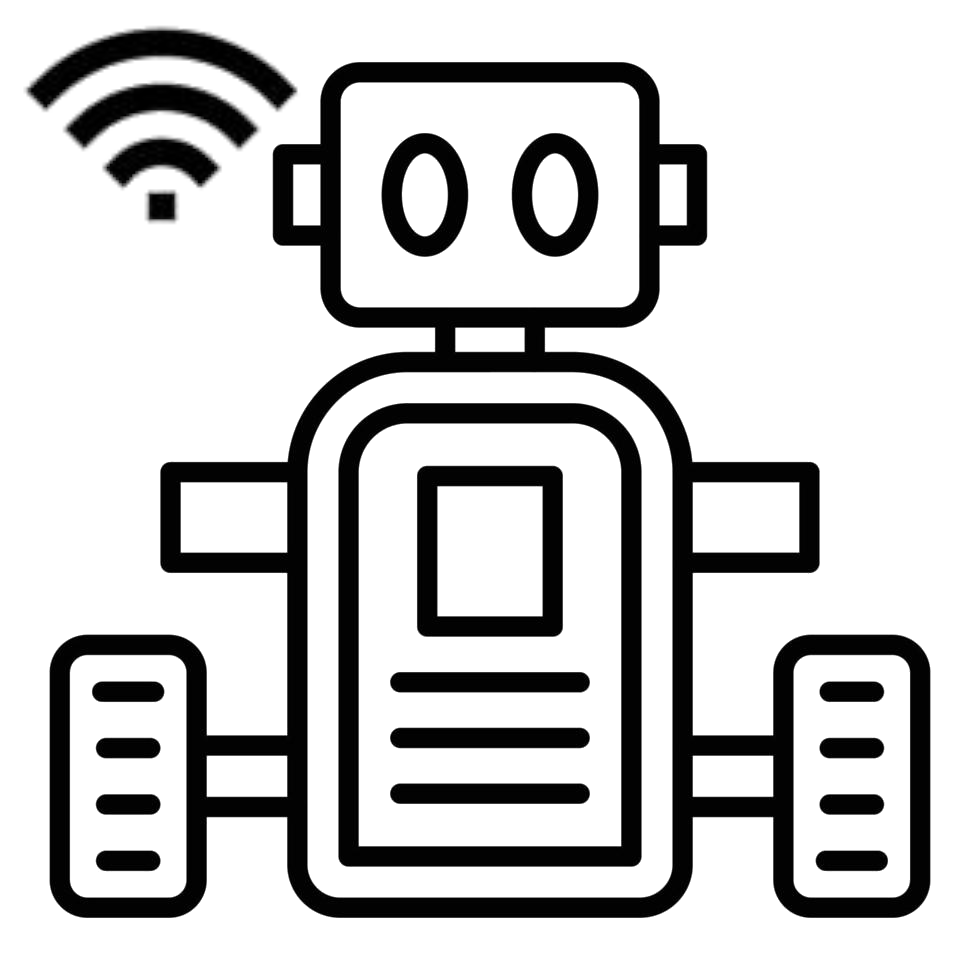}};
        \node[anchor=west, align=left] () at (Adam.south east){\textit{Adam}};
        \node[anchor=south, align=center, color=darkred] () at ([yshift=-0.1cm] Adam.north){\textbf{Risk Map}};

        \draw[dashed, color=darkblue, line width=1.5pt] ([xshift=0.3cm] Brange.center) circle (0.2cm);
        \draw[dashed, color=darkblue, line width=1pt] ([xshift=0.3cm] Brange.center) circle (0.4cm);
        \draw[dashed, color=darkblue, line width=0.5pt] ([xshift=0.3cm] Brange.center) circle (0.6cm);
        \node[] (Bob) at ([xshift=0.3cm] Brange.center) {\includegraphics[width=1cm]{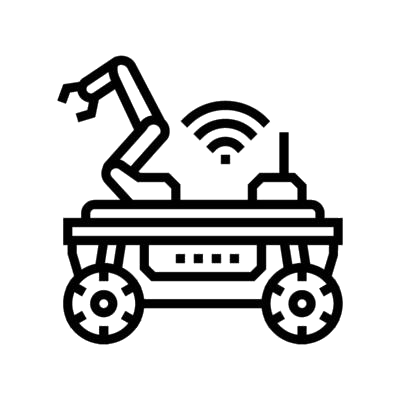}};
        \node[anchor=east, align=right] () at (Bob.north west){\textit{Bob}};
        \node[anchor=north, align=center, color=darkblue] () at ([yshift=0.1cm] Bob.south){\textbf{Risk Map}};

    \end{tikzpicture}
    \hspace{\fill}
    \caption{A trustworthy warehouse, where the robots communicate awareness (described by regions and risk maps).}
    \label{fig:scenario}
\end{figure}

A trustworthy \gls{mas} means all its components are aware of their situations and behave as expected~\cite{lewis2018role}.
According to~\cite{sifakis2023trustworthy}, a \textit{trustworthy} \gls{mas} should be able to analyze, mitigate, and evaluate risk in dangerous situations when agents interact with the environment.
As addressed in~\cite{ramchurn2004trust}, an agent in a trustworthy \gls{mas} will \textit{do exactly what it promises}. Consider a simple warehouse in Fig.~\ref{fig:scenario}, where two robots are required to move within the red and the blue regions, respectively, while avoiding collisions with each other.
To guarantee safety, each agent should carefully evaluate the risk of collisions based on the positions of each other and plan its motion accordingly.
If both robots promise to always remain in their domains, each robot only has to perform risk evaluation in the intersection region.
This may save resources and energy compared with a whole-space evaluation. Therefore, a trustworthy \gls{mas} with \gls{sa} can bring in higher efficiency and less consumption. On the other hand, it requires all agents to actively perform \gls{sa} and set up secure and reliable communication to exchange such awareness with other agents.

Developing secured systems with \gls{sa} is a popular topic in many fields, such as robotics~\cite{ginesi2020autonomous}, behavioral modeling~\cite{castaldo2015bayesian}, and \gls{adas} development~\cite{rockl2007architecture}.
Several tools have been developed over the years to aid other researchers in this effort, each focusing on some specific aspect of the simulation.
They can be generic, such as MESA~\cite{cite:mesa}, OpenMAS~\cite{cite:openmas} and MATLAB Simulink~\cite{cite:simulink},
proper physics simulators, like PyBullet~\cite{cite:pybullet} and Chrono~\cite{cite:chrono}
or concentrate on a more narrow field, such as Prescan~\cite{cite:prescan} and CARLA~\cite{cite:carla} in respect to autonomous driving.
Scenic~\cite{cite:scenic} fills another niche, focusing on the creation of complex multi-agent scenarios with a domain-specific language.
It uses simulators such as CARLA and GTA-V under the hood and it can enforce hard and soft constraints over the valid states a scenario can assume. Nevertheless, these existing tools still can not satisfy the growing demand for trustworthy \glspl{mas} due to the lack of specialized components for situation awareness and generic interfaces for knowledge and risk representation.

This paper introduces \symaware, a novel software framework designed to facilitate the development of trustworthy \glspl{mas} with \gls{sa}. This development framework provides a high-level abstract interface for an agent to understand its situation, represent its knowledge, and reason about its risk in an interactive environment. With a rich set of predefined components, each agent can efficiently compute its control input based on the awareness of its situation, knowledge, and risk and communicate with concerned agents. Meanwhile, \symaware can simplify the user's interaction with underlying simulators, such as Prescan\textregistered~and Pybullet by enabling a concurrency-based asynchronous mode. In this sense, \symaware creates a clear separation between system control and physical simulation. 
%Agents are assumed to have some form of \gls{sa} and knowledge they can leverage in their decision-making processes, as well as the ability to communicate with each other during the simulation.
Moreover, \symaware allows users to customize the controllers and simulation environments for the \gls{mas}, promising fine-grained control over the \gls{sa} of the whole system. Written in Python, a widely-used programming language, \symaware allows for seamless integration with a plethora of other well-established tools. Moreover, this framework is flexible in integrating various control and planning algorithms based on appropriate logical frameworks and symbolic computations, such as linear-time temporal logic (LTL) and signal temporal logic (STL).

\textit{Online Repository:} the source code of the \symaware framework and its documentation are hosted on GitLab~\cite{symaware2024framework}.

\section{Mathematical Models}

%This section elaborates the mathematical models of a \gls{mas} with \gls{sa}, formulating the foundation of \symaware.

\subsection{Control Based on Situational Awareness (SA)}

The \gls{sa} of each agent in a \gls{mas} represents its understanding of the state and intention of other agents and the uncertainty and risk of the environment. It is described by an \gls{sa} model and is regularly updated.
For a \gls{mas} consisting of $n$ agents, each single agent $i\in \mathcal{N}:= \{1,2,\cdots,n\}$ is described as a discrete-time dynamic model,
\begin{equation}\label{eq:am}
    x_{k+1}^{(i)} = f^{(i)}(x_k^{(1)}, \ldots, x_k^{(n)}, u_k^{(i)}, \epsilon_k^{(i)}),
\end{equation}
where $x_k^{(i)}\in \mathbb{X}$ and $u_k^{(i)} \in \mathbb{U}$ denote the state and the control input of the $i$-th agent at time $k\in \mathbb{Z}^+$, respectively, $\epsilon_k^{(i)}$ is random noise denoting the stochastic uncertainty of the agent, and $f^{(i)}$ is a function describing the physical dynamics of agent $i$. Due to the influence of the random noise, the trajectory of each agent $i$ till time $k$, denoted as $\mathbf{x}^{(i)}_k := x^{(i)}_0 x^{(i)}_1 \cdots x^{(i)}_k$, is a stochastic signal. When the system uncertainty is ignored, the random noise $\varepsilon^{i}_k$ can be omitted, leading to each agent a deterministic trajectory $\mathbf{x}^{(i)}_k$. Eq.~\eqref{eq:am} assumes all agents are fully connected since its update depends on the states of all agents. More common is a partially connected \gls{mas} for which Eq.~\eqref{eq:am} only depends on the states of the neighbors of agent $i$~\cite{sun2018event}.

For such a \gls{mas}, \gls{sa} at a discrete time $k\in \mathbf{Z}^+$ is described as $\sigma^{(i)}_k:=\{b^{(i)}_k,\,\iota^{(i)}_k,\,v^{(i)}_k,\,r^{(i)}_k\}$ where $b^{(i)}_k$, $\iota^{(i)}_k$, $v^{(i)}_k$, and $r^{(i)}_k$ are the belief state, intent, quantified uncertainty, and risk of agent $i$ at time $k$, respectively. We can omit $v^{(i)}_k$ and let $b_k^{(i)}:=x_k^{(i)}$ for a deterministic \gls{mas}. $\sigma^{(i)}_k$ can be inferred from the historical trajectory~\cite{stroeve2007safety}.  The \gls{sa} of each agent is updated by the following dynamic model~\cite{blom2015modelling},
\begin{equation}\label{eq:sa}
    \sigma^{(i)}_{k+1} := f^{(i)}_{\mathrm{SA}}(\sigma^{(1)}_k, \cdots, \sigma^{(n)}_k , u^{(i)}_k, \mathbf{x}^{(1)}_k, \ldots, \mathbf{x}^{(n)}_k),
\end{equation}
where $f^{(i)}_{\mathrm{SA}}$ is an updating function to be designed. The objective of an \textbf{SA-based control problem} is: \textit{to design an SA-updating function $f^{(i)}_{\mathrm{SA}}$ and a controller $u_{k}^{(i)}$ for agent $i$, such that its infinite trajectory $x_0^{(i)}x_1^{(i)}\cdots$ satisfies a predefined specification which will be introduced as follows.}

%The updating of the \gls{sa} model in Eq.~\eqref{eq:sa} involves the following three stages.
% \begin{itemize}
%     \item \textbf{Reasoning:} the \gls{sa} $\sigma^{(i)}_k$ is calculated without considering any interactions with other agents;
%     \item \textbf{Observation:} the \gls{sa} $\sigma^{(i)}_k$ is updated considering a unidirectional information flow from other agents;
%     \item \textbf{Communication:} the \gls{sa} is updated  
% \end{itemize}

\subsection{Control Based on Risk Awareness}\label{sec:ra}

Risk refers to the possibility of failing a predefined specification. The specification for a \gls{mas} can be described as a temporal logic formula expressing a rich class of practical tasks like goal-reaching and collision avoidance. In this sense, risk awareness refers to the process of quantifying and restricting the probability of failing a specification. \gls{ltl} is a widely used fragment of temporal logic with the following inductively defined syntax,
\begin{equation}\label{eq:ltl}
    \varphi::= p \mid \lnot \varphi \mid \varphi_1 \wedge \varphi_2 \mid \varphi_1 \vee \varphi_2 \mid \bigcirc \varphi \mid \varphi_1 \mathsf{U} \varphi_2,
\end{equation}
where $p \in \mathsf{AP}$, with $\mathsf{AP}$ being a finite set of atomic propositions describing the desired properties of a \gls{mas}, $\lnot$, $\wedge$, $\vee$, $\bigcirc$, and $\mathsf{U}$ are the \textit{not}, \textit{conjunction}, \textit{disjunction}, \textit{next}, and \textit{until} operators, respectively. Let $\Sigma:=2^{\mathsf{AP}}$ be an alphabet generated by $\mathsf{AP}$, and let $\mathcal{L}: \mathbb{X} \rightarrow \Sigma$ be a labeling function. Then, any run of the \gls{mas} in Eq.~\eqref{eq:am} generates a word $\mathcal{L}(x_0^{(i)})\mathcal{L}(x_1^{(i)})\cdots$, for $i\in \mathcal{N}$. Let $w_k=\mathcal{L}(x_t^{(i)})\mathcal{L}(x_{t+1}^{(i)})\cdots$ be its suffix of this word for any $t \in \mathbb{N}$. The satisfaction of an LTL formula $\varphi$ by a specific path $w_k$ is denoted by $w_k \models \varphi$ and defined recursively as 
$w_k \models p \Leftrightarrow p \in \mathcal{L}(x_k^{(i)})$,
$w_k \models \lnot \varphi \Leftrightarrow \lnot (w_k \models \varphi)$,
$w_k \models \varphi_1 \wedge \varphi_2 \Leftrightarrow w_k \models \varphi_1 \wedge w_k \models \varphi_2$,
$w_k \models \varphi_1 \vee \varphi_2 \Leftrightarrow w_k \models \varphi_1 \vee w_k \models \varphi_2$,
$w_k \models \bigcirc \varphi \Leftrightarrow w_{k+1} \models \varphi$,
$w_k \!\models\! \varphi_1 \mathsf{U} \varphi_2 \Leftrightarrow w_{k+i} \!\models\! \varphi_2,\exists\,i \!\in\! \mathbb{N}~ \mathrm{and}~w_{k+j} \!\models\! \varphi_1,\, \forall j \!\in\! \mathbb{N},\, j\! <\! i$.

In many practical applications, signal temporal logic (STL) is commonly used. STL has a similar definition of syntax and semantics to LTL but is defined on real-valued signals. The readers may refer to~\cite{Maler2004} for the details of STL.

With the system uncertainty, the risk is quantified as the probability that a predefined specification $\varphi$ fails, i.e., $P(w_k \nvDash \varphi)$. Therefore, the objective of a \textbf{risk-aware control problem} can be formulated as: \textit{to design a controller for the \gls{mas}, such that this path $w_k$ satisfies $P(w_k \models \varphi) > 1 - \varepsilon$, where $\varepsilon \in [\,0,\,1\,]$ denotes the predefined maximal risk level.}

\subsection{Control Based on Knowledge Awareness}

Knowledge describes how an agent should interact with the environment. The efficiency of decision-making and risk awareness of a \gls{mas} can be promoted by understanding and utilizing the knowledge about the system and the environment. Such a process has been referred to as \textit{knowledge awareness}~\cite{calvagna2023using} which enables the system to make informed decisions with enhanced safety. For physical dynamic systems, knowledge can be represented as temporal logic specifications~\cite{bacchus2000using} defined in Eq.~\eqref{eq:ltl} for an abstract model of the physical system in Eq.~\eqref{eq:am}. Consider a set of atomic propositions $\mathsf{AP}$ as mentioned in Sec.~\ref{sec:ra}. The abstraction model of the system in Eq.~\eqref{eq:sa} is a \gls{fts} which is a tuple $T:=\{Q, Q_0, \mathcal{F}, O, \mathcal{O}, T\}$, where $Q$ is a finite set of states, $Q_0 \subset Q$ represents the set of initial states, $\mathcal{F}: Q \rightarrow 2^{Q}$ denotes state transitions, $O$ is the set of observations, $\mathcal{O}: Q \rightarrow O$ is the mapping of observation, and $\mathcal{L}_T: Q \rightarrow 2^{\mathsf{AP}}$ is a labeling function that associates desired properties with the states. A bisimilar \gls{fts} for Eq.~\eqref{eq:sa} ensures they always generate the same word when $x_0^{(i)} \in Q_0$.

A fragment of \gls{ltl} named \textit{Generalized Reactivity} (GR(1)) is commonly used to specify the interaction between a system and the environment. It can be represented as the following conjunctive normal form (CNF),
\begin{equation*}
    \textstyle    \varphi_{\mathrm{I}}^{(e)} \wedge \bigwedge_i \square \varphi_{\mathrm{S},i}^{(e)} \wedge \bigwedge_j \square \lozenge \varphi_{\mathrm{F},j}^{(e)}  \Rightarrow \varphi_{\mathrm{I}}^{(s)} \wedge \bigwedge_i \square \varphi_{\mathrm{S},i}^{(s)} \wedge \bigwedge_j \square \lozenge \varphi_{\mathrm{F},j}^{(s)},
\end{equation*}
where formulas with a superscript $e$ denote the ones encoding the assumptions on the environment and those with $s$ specify the system. Meanwhile, the formulas with subscripts $\mathrm{I}$, $\mathrm{S}$, and $\mathrm{F}$ are Boolean formulas encoding initial states, safety, and liveness goals, respectively. Then, the \textbf{knowledge-aware control problem} can be formulated as: \textit{given the model of the system as a \gls{fts} $T$ and a GR(1) specification $\varphi$ to encode the knowledge about the environment, to design a reactive controller automatically to satisfy the specification.}

% \subsection{Maintaining the Integrity of the Specifications}

\section{SymAware Architecture}

\begin{figure}[htbp]
    \noindent
    \hspace*{\fill}
    \begin{tikzpicture}[scale=1, font=\small]

        \definecolor{sorange}{RGB}{255, 205, 153}
        \definecolor{sblue}{RGB}{204, 221, 255}
        \definecolor{sred}{RGB}{255, 235, 229}
        \definecolor{sgreen}{RGB}{229, 255, 229}
        \definecolor{syellow}{RGB}{255, 255, 245}
        \definecolor{spurple}{RGB}{255, 229, 255}
        \definecolor{sgray}{RGB}{242, 242, 242}

        \node[minimum width=8.2cm, minimum height=6.2cm, rounded corners=0.2cm, draw, dashed, fill=syellow] (fw) at (2.4cm, 0.4cm) {};
        \node[align=left, anchor=south west] () at ([xshift=-1.6cm, yshift=1.6cm] fw.south east) {\large \textbf{Agent}};

        \node[minimum width=3.0cm, minimum height=5.8cm, text width=1.8cm, draw,thick, fill=sred] (ag) at (0cm, 0.4cm) {};

        \node[minimum width=2.4cm, minimum height=1.2cm, text width=1.8cm, draw, align=center, dashed] (kw) at (0cm, 2.4cm) {\textbf{Knowledge Awareness}};

        \node[minimum width=2.4cm, minimum height=2.8cm, text width=2cm, draw, align=center, dashed] (sa) at (0cm, 0.2cm) {\textbf{Situational Awareness} \\ \vspace{0.2cm}
            \textit{State} \\
            \textit{Intent} \\
            \textit{Uncertainty} \\
            \textit{Risk}};

        \node[minimum width=2.4cm, minimum height=0.8cm, text width=2.2cm, draw, align=center,thick,dashed] (ru) at (0cm, -1.8cm) {\textbf{Risk Awareness}};

        \node[minimum width=4.2cm, minimum height=0.6cm, draw, align=center,thick,fill=spurple] (uc) at (4.2cm, -2.2cm) {\textbf{Uncertainty Component}};

        %\draw [<->, >=Stealth, thick] (ag) -- (ru);

        \draw [<->, >=Stealth, thick] ([xshift=-0.6cm] uc.west) -- (uc.west);

        \node[minimum width=2.6cm, minimum height=0.6cm, text width=2.4cm, draw, align=center,thick, fill=sgreen] (pc) at (3.4cm, 2.9cm) {\textbf{Perception}};

        \draw [->, >=Stealth, thick] (pc.west) -- ([xshift=-0.55cm] pc.west);

        \draw [<-, >=Stealth, thick] (pc.east) -- ([xshift=0.5cm] pc.east);
        \fill[black] ([xshift=0.5cm] pc.east) circle (2pt);

        \node[minimum width=2.6cm, minimum height=0.6cm, text width=2.4cm, draw, align=center,thick, fill=sblue] (em) at (3.4cm, 2.1cm) {\textbf{Entity - Model}};

        \draw [<-, >=Stealth, thick] (em.west) -- ([xshift=-0.55cm] em.west);

        \draw [->, >=Stealth, thick] (em.east) -- ([xshift=0.5cm] em.east);
        \fill[black] ([xshift=0.5cm] em.east) circle (2pt);

        \node[minimum width=2.6cm, minimum height=0.8cm, text width=2.4cm, draw, align=center,thick, fill=sorange] (cs) at (3.4cm, 1cm) {\textbf{Communication Sender}};

        \draw [<-, >=Stealth, thick] (cs.west) -- ([xshift=-0.55cm] cs.west);

        \draw [->, >=Stealth, thick] (cs.east) -- ([xshift=0.5cm] cs.east);
        \fill[black] ([xshift=0.5cm] cs.east) circle (2pt);

        \node[minimum width=2.6cm, minimum height=0.8cm, text width=2.4cm, draw, align=center,thick, fill=sorange] (cr) at (3.4cm, 0cm) {\textbf{Communication Receiver}};

        \draw [->, >=Stealth, thick] (cr.west) -- ([xshift=-0.55cm] cr.west);

        \draw [<-, >=Stealth, thick] (cr.east) -- ([xshift=0.5cm] cr.east);
        \fill[black] ([xshift=0.5cm] cr.east) circle (2pt);

        \node[minimum width=4.2cm, minimum height=0.6cm, text width=4cm, draw, align=center,thick,fill=sgray] (cc) at (4.2cm, -1.4cm) {\textbf{Controller Componet}};

        \draw [<->, >=Stealth, thick] (cc.west) -- ([xshift=-0.55cm] cc.west);

        \node[minimum width=1.4cm, minimum height=1.8cm, text width=1.2cm, draw, dashed, align=center, fill=white] (env) at (6cm, 2.5cm) {\textbf{Environ- ment}};

        \node[minimum width=1.4cm, minimum height=1.8cm, text width=1.2cm, draw, dashed, align=center, fill=white] (oa) at (6cm, 0.5cm) {\textbf{Other Agents}};

    \end{tikzpicture}
    \hspace{\fill}
    \caption{\symaware's architecture conceptualized initially in~\cite{kordabad2024robust}.}
    \label{fig:symaware-implementation}
\end{figure}
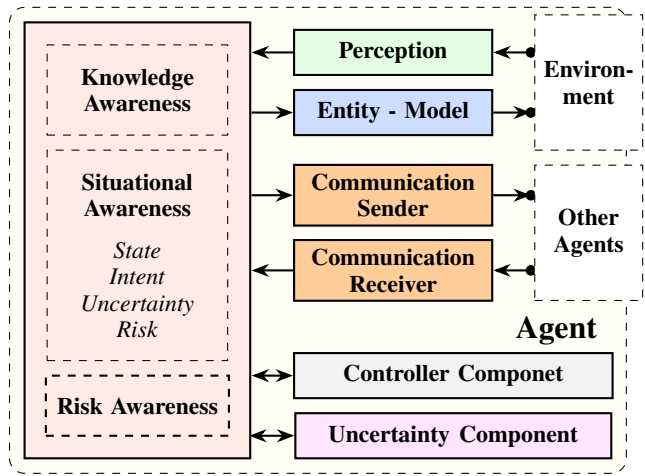

\symaware aims to create an easy-to-use tool to develop \gls{sa}-enabled \gls{mas} for less experienced developers. It is extendable for general scenarios with minimal adjustments. Moreover, it highlights the capability of modularization, allowing users to flexibly customize components according to specific requirements. Fig.~\ref{fig:symaware-implementation} illustrates the modularized conceptual architecture of \symaware.

\subsection{Agent}
To achieve a high degree of modularity, the \symaware framework is built around the interaction between agents and their underlying components following an \gls{oop} design paradigm. \gls{oop} addresses an agent and its specialized components.
% To achieve the desired modularity, the framework is built around the concept of agents and components interacting with each other.
% As long as the shared interface is respected, the user can easily create new components and agents, or swap out existing ones.
% \gls{oop} naturally fits this design, as it allows for the creation of classes that can be easily extended and overridden.
%
%
An agent is a collection of multiple components having all access to the same, shared state.
The state can be divided into two main parts: the agent's awareness and the knowledge database.
The agent's awareness contains information about the agent's internal state, (e.g. its position, velocity, and acceleration), intent, risk, and uncertainty associated with the agent's actions.
The knowledge database is a more flexible key-valued data storage applied to store higher-level information (e.g. maps of the environment, high-level tasks), needed by the agent's components for their computations.
Multiple agents can be instantiated in the same simulation allowing for the creation of complex multi-agent systems.
To interact with the environment, each agent is associated with an entity, which in turn is assigned a model (see \ref{sec:environment-entity-models}).

\subsection{Components}

Components are functions to update the \gls{sa} of an agent, ensuring the flexibility and modularization of \symaware by allowing users to create customized component functions.
Every concrete component must implement the abstract protected methods \texttt{\_compute} and \texttt{\_update}.
No other alteration to the base class is required, although it is possible to introduce new utility methods or fields in the subclasses or to override the base class methods if needed, although it is advisable to call the corresponding super class' method to ensure the expected behavior of the component.
The \texttt{\_compute} method is responsible for processing the information available to the agent and producing a new value.
While there is no limitation on the number of arguments it can take, it is recommended to keep its signature as simple as possible, since most information can be collected from the agent’s internal state directly.
The \texttt{\_update} method is responsible for updating the agent’s internal state based on the result of the computation.
It must accept the value computed by \texttt{\_compute} as an argument.
How the value is used and what changes are applied to the agent’s internal state is up to the component’s implementation.

\begin{listing}[ht]
    \begin{minted}[fontsize=\footnotesize]{python}
class MyController(Controller):
  def _compute(self):
    state = self._agent.self_state
    goal = self._agent.self_knowledge["goal"]
    control_input = goal - state
    return control_input, TimeSeries()

  def _update(self, value):
    control_input, intent = value
    self._agent.model.control_input = control_input
    self._agent.self_awareness.intent = intent
\end{minted}
    \caption{Example of a customized controller component. It computes its control input based on the states and intents of agents of interest and its knowledge of the goal position.}
    \label{lst:controller}
\end{listing}

\symaware provides a set of specialized components (Fig. \ref{fig:symaware-implementation}) that provide some built-in utility methods, covering the most common use cases.
These components can be further extended or overridden to fit the user's specific requirements.

\subsubsection{Perception component}

The Perception System can act as a limited sensor or as an omniscient oracle responsible for querying the environment and for the current state of the agent or any other entity of interest.

\subsubsection{Risk and Uncertainty components}
The knowledge database and \gls{sa} vector will be used by the risk and uncertainty components to compute and store new values for the risk and uncertainty to consider in the control computation.

\subsubsection{Communication components}
Optionally, the agent can use the communication components to share some information with other agents, helping them in their decision-making process.
The communication components can support a variety of communication protocols (e.g. message queue, TCP).

\subsubsection{Controller component}
Often the last step in the simulation loop involves feeding the update agent’s state to the controller.
Its role is to produce the control input to apply to the model, as well as an intent that will be stored in the \gls{sa} and could play a role in successive steps.
The control input applied to the model will then be played out in the environment, and the new state will be available to the component in the next iteration.
The controller is usually the most important component to implement, as it is the one that, in the end, will determine the agent’s behavior in the environment.
The shape and type of control input the controller will return at the end of the \texttt{\_compute} method depends on the agent’s model, which in turn depends on the environment and the simulator used (see \ref{sec:environment-entity-models}).

\subsection{Environment, Entity, and Models}
\label{sec:environment-entity-models}
\symaware uses environment, entity, and model classes as in the Facade design pattern \cite{cite:design-pattern} for generalization.
While the exposed \gls{api} should be deceptively simple and easy to use, the underlying implementation may be quite complex.
Using this approach, once an environment and related entities and models have been implemented, users can utilize them without having to worry about low-level details.
This makes it easy to switch between simulators with very few changes required.
\symaware currently provides basic support for PyBullet \cite{cite:pybullet}, Prescan\textregistered~\cite{cite:prescan}, and Pymunk \cite{cite:pymunk}, but it is easy to extend the existing models or add support for other simulators.

\subsubsection{Environment}
The environment in \symaware is an interface that abstracts away the simulator's specifics.
It handles the setup of the simulation, the instantiation of the entities, and the execution of the simulation step, as well as the optional cleanup.
At each step, the simulator will execute the step function of all registered entities.
If the entity has an associated model, its step function will provide the underlying simulator with the information it needs to update the entity's state (e.g. the force applied to the entity).

\subsubsection{Entity}
Entities are the physical objects that inhabit the environment.
Every agent must be associated with an entity to be able to interact with other agents and the environment.
On the other hand, entities can exist without agents to represent generic obstacles.
% An entity without a model won't receive any inputs from the framework, but it may still move due to the effect of an external force on it (e.g. a collision with some other entity, gravity).

\subsubsection{Model}
The model represents the entity's dynamics applied over the entity's state at every simulation step based on the control input provided by the agent's controller or inputs from the environment's underlying implementation.

\section{SymAware Mechanism}
This section interprets the mechanism of
\symaware. Two different modes of operation are allowed, namely the \textit{Synchronous mode}, where all components run in a predictable sequence, and the \textit{Asynchronous mode}, where concurrency between the components is allowed. \symaware also provides an event-triggered mechanism, where responsive functions can be called upon certain actions. A simple illustration of the \symaware operation mechanism is given in Fig.~\ref{dg:main-loop}.

    \begin{figure}[h]
        \begin{adjustbox}{width=.4\textwidth,center}
            \input{fig/main-loop}
        \end{adjustbox}
        \caption{\small
    The simulation mechanism of \symaware, showcasing how the controller component of an agent updates its state.
    Events are triggered Before and after each updating step, allowing adding custom behavior via subscription.
}\label{dg:main-loop}
    \end{figure}

\subsubsection*{Synchronous mode}
In this mode, the agent calls the \texttt{compute\_and\_update} method for each of its components. Such a process is repeated for all agents by an agent coordinator. Although simple for implementation and debugging, this mode may suffer from slow processing and long execution time due to busy waiting.

\subsubsection*{Asynchronous mode}
If the simulation is running the asynchronous mode, each component starts a new awaitable routine that runs the \texttt{async\_compute\_and\_update} method in a loop, with a lock between each iteration determining the frequency at which the component will run.
The lock may be time-based or event-based.
By default, the async methods will still call the \texttt{\_compute} and \texttt{\_update} abstract methods, but
there is the option to override their asynchronous counterpars, \texttt{\_async\_compute} and \texttt{\_async\_update} to leverage the \textit{asyncio} library.
While the asynchronous approach is only concurrent by default, it can easily be integrated with multiprocessing to achieve true parallelism.

    \begin{figure}[h]
        \begin{adjustbox}{width=.35\textwidth,center}
            \input{fig/asynchronous}
        \end{adjustbox}
        \caption{The asynchronous mode, where a context is switched when a coroutine is called by the \textit{await} keyword, allowing other components to run without waiting for a long operation or loop.}\label{dg:asynchronous}
    \end{figure}

\subsubsection*{Events}
In every iteration, components emit events that can be associated with customized callback functions.
The default implementation triggers an event before and after every important action in the simulation loop, such as before and after \texttt{\_compute}, \texttt{\_update} and \texttt{initialize} method calls, forwarding to the user the agent and any relevant data.
The user can exploit this feature to add their own custom behavior to the component without modifying it internally. \par

\section{Case Studies}

In this section, we use three \gls{mas} control studies to validate the efficacy and efficiency of the \symaware framework. Each study covers the awareness of \textit{intent}, \textit{goal}, and \textit{risk} to facilitate the SA of a \gls{mas}. The details of these case studies can be found in~\cite{cite:githubrepo}.

\subsection{Use Case I: Intent-Aware Collision Avoidance}

\begin{figure}[htbp]
    \centering
    \begin{subfigure}{0.4\linewidth}
\includegraphics[width=0.9\columnwidth]{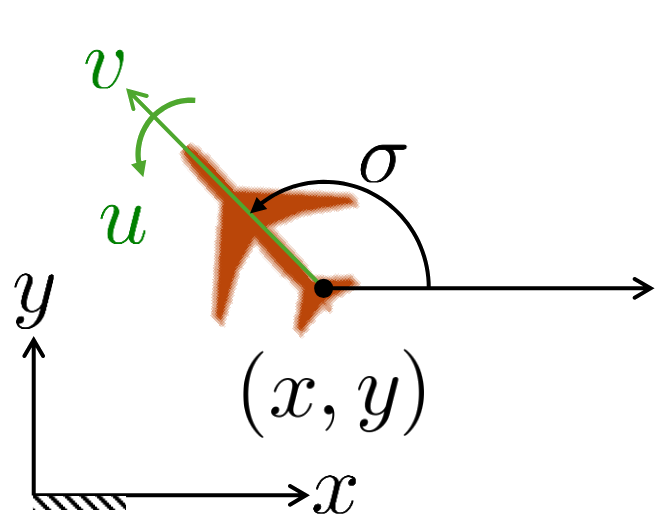}
    \caption{}
    \label{Fig:aircraft}
\end{subfigure}
    \begin{subfigure}{0.55\linewidth}
\includegraphics[width=0.95\columnwidth,trim={6cm 11cm 0 0},clip]{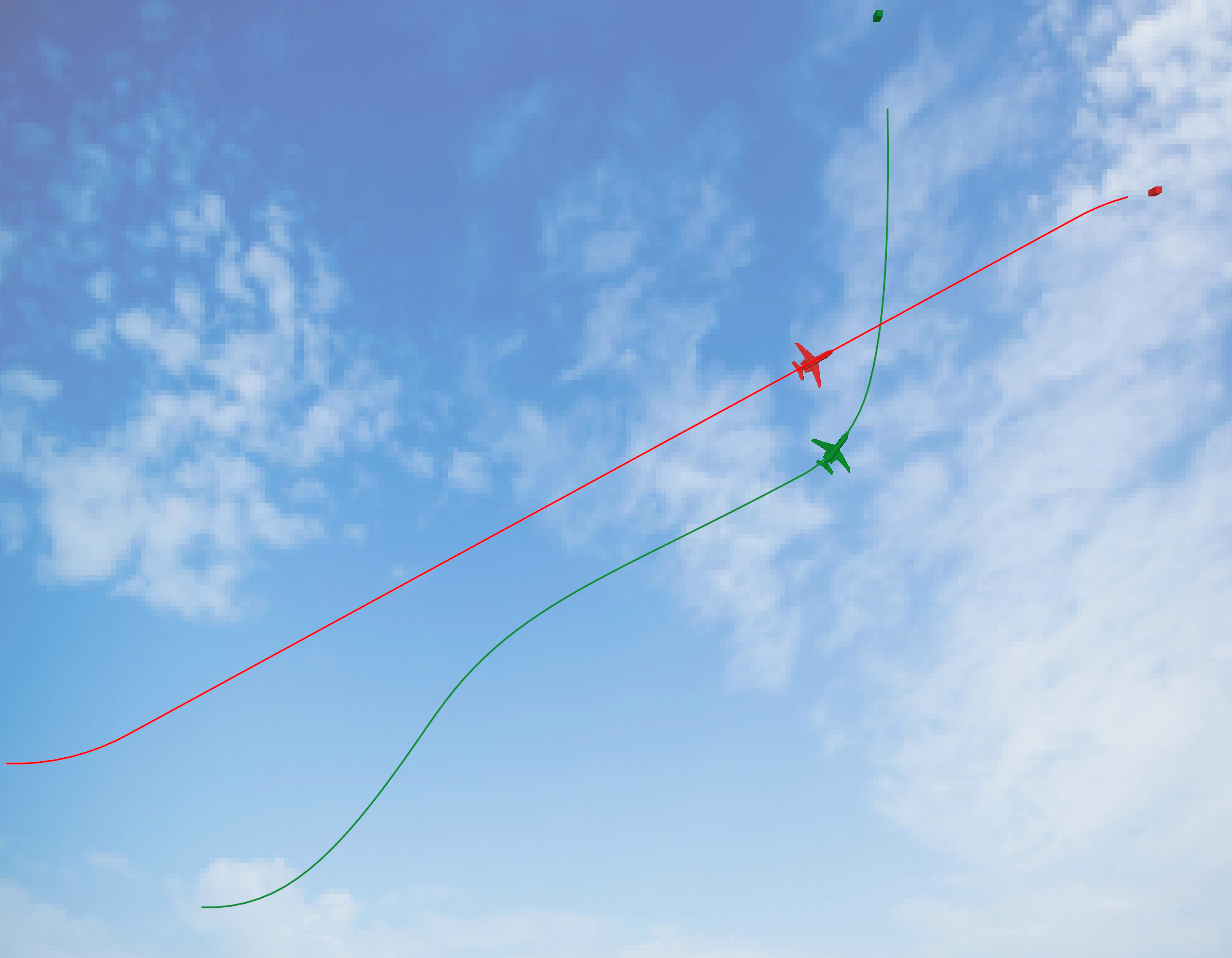}
    \caption{}
    \label{Fig:sim3}
\end{subfigure}
    
    \caption{
        (a). the aircraft in the horizontal plane in the earth-fixed coordinate, where black variables are the state and the greens are the inputs. (b). The ownership (green) avoids the intruder (red) while both move toward their destinations.}
\end{figure}

Consider two aircraft, an ownership, and an intruder, with the following equation of motion in the discrete-time setting:
\begin{equation}\label{eq:single agent EoMs1}
    \vect s_{k+1}:=\begin{bmatrix}
        {x}_{k+1} \\ {y}_{k+1} \\ {\sigma}_{k+1}
    \end{bmatrix} = \begin{bmatrix}
        {x}_{k} \\ {y}_{k} \\ {\sigma}_{k}
    \end{bmatrix}+t_e
    \begin{bmatrix}
        v_k \cos{\sigma_{k}} \\ v_k \sin{\sigma_{k}} \\ u_k
    \end{bmatrix},
\end{equation}
where $\vect s_k$ is the state of the aircraft at time $k$, ${x}_{k}$ and ${y}_{k}$ are the position states, and ${\sigma}_{k}$ is the heading angle.
The constant $t_e$ is the sampling time, assumed to be $1$ sec.
$v_k$ and $u_k$ are the linear and angular velocities, respectively.
Figure~\ref{Fig:aircraft} shows the geometry of the problem. Both aircraft have a target position to reach, and the ownership has to avoid a collision with the intruder using a \gls{mpc} approach.
Instead of considering only the deterministic dynamics at play, the system's state at a certain step also considers $m$ samples from a distribution $W$.
The number of trajectories defined by this scenario grows exponentially with the length of the \gls{mpc} horizon.
Therefore, it is reasonable to fix the uncertain parameters after a certain period of time called the robust horizon $N_r < N$.
Hence, the number of scenarios is reduced to $M = m^{N_r}$.

We focus on scenarios where the ownership is aware of the intruder’s intent, which is assumed to be the Dublin path that leads the intruder to its destination \cite{cite:dublin}.
Note that, the linear velocity is assumed to take its maximum value (constant) for optimality purposes.
By incorporating the intruder’s intent information in the \gls{mpc} scheme, the robustness of the whole system is increased~\cite{kordabad2024robustmodelpredictivecontrol}.
The expression of the intent-aware cost function when the ownership is at the state $\vect s_t$, $0\leq t\leq T$, along with its constraints, is given by:
\begin{subequations}\label{eq:RMPC}
    \begin{align}\label{eq:RMPC:cost}
         & \textstyle \min_{\bm{u}} \,\,\,\,  \|\vect s^1_{N}-\vect s_T\|_{Q_{\mathrm{f}}}+\|\vect s^1_{0}-\vect s_T\|_{Q}  
         +\sum_{k=1}^{N-1}  \|\vect s^1_{k}-\vect s_T\|_{Q}\\
         &\;\quad \mathrm{s.t.} \,\,  \forall j\in\{1,\ldots, M\},  \forall k\in\{0,\ldots, N-1\}:\nonumber \\& \qquad \qquad 
         \eqref{eq:single agent EoMs1},\forall i\in\{1,\{2,j\}\}, \label{eq:RMPC:dyn}\\
         & \qquad\quad\,\,\, \textstyle \rho\leq \sqrt{(\vect s^1_{k}-\vect s^{2,j}_{k})^\top \mathrm{diag}(1,1,0) R(\vect s^1_{k}-\vect s^{2,j}_{k})},\label{eq:RMPC:safe} \\
         & \qquad\quad\,\,\,  \underline{u}^{1}\leq u^{1}_k\leq \bar{u}^{1},\quad \underline{v}^{1}\leq v^{1}_k\leq \bar{v}^{1},\label{eq:RMPC:own}             \\
         & \qquad\quad\,\,\, \text{if}\,\, k< N_r:\,\,\, u^{2,j}_k= \label{eq:RMPC:intr11}                                                                      \\ &\qquad\quad\,\,\, \left\{\begin{matrix} \bar{u}^{2}                    & \text{if}\, \mathrm{mod}(\left \lceil \frac{j}{3^{N_r-k-1}} \right \rceil,3)=0\, \\
               \underline{u}^{2}              & \text{if}\, \mathrm{mod}(\left \lceil \frac{j}{3^{N_r-k-1}} \right \rceil,3)=1\, \\
               D(\vect s^2_0,\vect s^2_T,t+k) & \text{if}\, \mathrm{mod}(\left \lceil \frac{j}{3^{N_r-k-1}} \right \rceil,3)=2,
        \end{matrix}\right.\nonumber\\
         & \qquad\quad\,\,\, \text{if}\,\, k\geq N_r:\,\,\, u^{2,j}_k=D(\vect s^2_0,\vect s^2_T,t+k),\label{eq:RMPC:intr12}                                     \\
         & \qquad\quad\,\,\, v^{2,j}_k=\bar{v}^{2}, \quad \vect s^1_0=\vect s_t \label{eq:RMPC:intr2}
    \end{align}
\end{subequations}
where $\bm{u} = [u_{0:N-1}^1,v_{0:N-1}^1,\vect s_{0:N}^1]$, $N$ is the horizon length, $Q_{\mathrm{f}}$ and $Q$ are positive definite matrices, $R$ is a positive constant, $\rho$ is the minimum allowed horizontal distance of the crafts, and $\underline{v}^{i}$ ($\underline{u}^{i}$) is the minimum and $\bar{v}^{i}$ ($\bar{u}^{i}$) is the maximum linear (angular) velocity for $i\in\{1,2\}$.
The function $D(\vect s^2_0,\vect s^2_T,t+k)$ yields the waypoint on the Dublin path at time $t+k$.
% \begin{figure}[b]
%     \includegraphics[width=0.9\columnwidth,trim={6cm 11cm 0 0},clip]{case3.jpeg}
%     \centering
%     \caption{
%         The ownship (green) avoids the intruder (red) while both move towards their destinations.
%     }
%     \label{Fig:sim3}
% \end{figure}

\subsection{Use Case II: Multi-Agents Spatio-Temporal Goals}
Consider a team of agents with indices $i$ in $\mathcal{V} = \{1, \ldots 5\}$ and a set of edges $\mathcal{E}_{\psi} \subset \mathcal{V}\times \mathcal{V}$ termed as the \textit{task edges}, such that $\mathcal{N}(i) :=\{j\in \mathcal{V} | (i,j) \in \mathcal{E}_{\psi} \land j\neq i\}$ is the set of neighbours of agent $i$. We let $x_k^i \in \mathbb{R}^n$ and  $u_k^i \in \mathbb{R}^m$ be the state and control input of agent $i$ at time $t_k$, while $x^i(t)$ and $u^i(t)$ represent the continuous time state and control input. Each agent follows the continuous and sampled data dynamics:
\begin{equation} \label{eq:system dynamics case 2}
    \begin{aligned}
        \dot{x}^i(t) = f^i_c(x^i(t)) + g^i_c(x^i(t))u^i(t),\;\;
        x_{k+1}^i = f^i(x_k^i) + g^i(x_k^i)u_k^i,
    \end{aligned}
\end{equation}
where functions $g^i_c : \mathbb{R}^{n\times m} \rightarrow \mathbb{R}^n,\; f^i_c : \mathbb{R}^n \rightarrow \mathbb{R}^n$ represent the continuous-time dynamics, while $g^i : \mathbb{R}^{n\times m} \rightarrow \mathbb{R}^n,\; f^i : \mathbb{R}^n \rightarrow \mathbb{R}^n$ represent the dynamics under zero order-hold sampling for agent $i$. We denote by $x_k = [x_k^i]_{i\in \mathcal{V}} \in \mathbb{R}^{|\mathcal{V}|n}$ the global state of the team at time $t_k$, obtained by vertically stacking the state of every single agent and we consider the case a spatio-temporal goal expressed in the Signal Temporal Logic formalism as $\psi : = (\bigwedge_{i\in \mathcal{V}}  \phi_{i}) \land (\; \bigwedge_{(i,j)\in \mathcal{E}_{\psi}}\;  \phi_{ij}))$ is assigned to the system. Namely, $\phi_{i}$ represents an independent goal for each agent, like visiting an interest point location and $\phi_{ij}$ represents a collaborative goal among a pair of agents $(i,j)\in \mathcal{E}_{\psi}$ like a relative formation to be achieved within some time constraints or a maximum distance constraint. Only the information about the local task $\phi_{i}$ and $\phi_{ij}, \;\forall j\in \mathcal{N}_{\psi}(i)$, is stored in the knowledge database of each agent without global knowledge of the global goal $\psi$.
\par
\begin{figure}
\vspace{0.3cm}
    \centering
    \includegraphics[width=\linewidth,trim={0 8cm 0 0},clip]{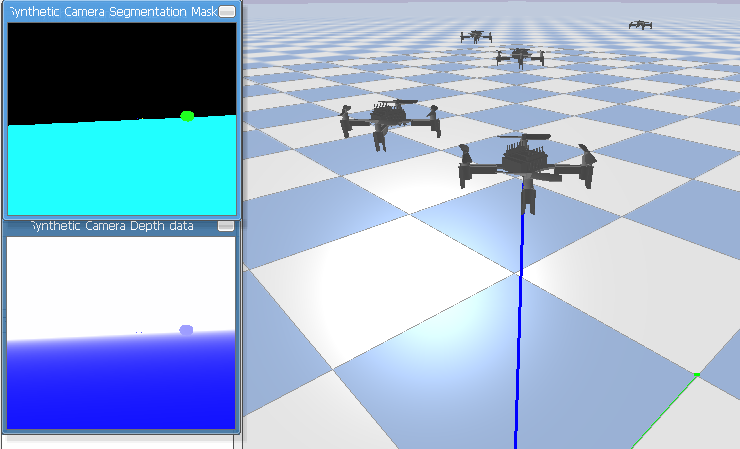}
    \caption{A team of drones in Pybullet with a camera directing the point of interest (sphere shape in the camera view on the left)}.
    \label{fig:drones}
\end{figure}

From the high-level specifications $\phi_{ij}$ and $\phi_{i}$, continuous and piece-wise differentiable real-valued Sampled-Data Control Barrier Functions $b_{\phi}^{ij} : (x^i,x^j,t) \mapsto \mathbb{R}$ and $b_{\phi}^{i} : (x^i,t) \mapsto \mathbb{R}$ can be programmatically constructed together with a pair of function $\nu^{ij}_{\phi}(x^i,x^j,t) : \mathbb{R}^{n} \times\mathbb{R}^{n} \times \mathbb{R}_{+} \rightarrow \mathbb{R} $ and  $\nu^{i}_{\phi}(x^i,t) : \mathbb{R}^{n} \times \mathbb{R}_{+} \rightarrow \mathbb{R}$ applied to extend the definition Control Barrier Functions to sampled data setting \cite{larsmain,gregcdc}. Let a state trajectory for the system be defined as $\bm{x} :=\{ x_k\}_{k\in 0,\ldots \infty}$, and let $\bm{x} \models \psi$ denote that the state trajectory $\bm{x}$ for the system satisfies the global task $\psi$. Moreover, let the graph $\mathcal{G}(\mathcal{V},\mathcal{E}_{\psi})$, induced by the vertices $\mathcal{V}$ and task links $\mathcal{E}_{\psi}$, be connected and acyclic, and let each agent $i$ be elected leader of a single edge $(i,r) \in \mathcal{E}_{\psi}$. The following decentralized control law to satisfy the local task $\phi_{i} \land \phi_{ir}$ is then applied:
\begin{subequations}
    \label{eq:new decentralized optimization with tasks}
    \begin{align}
         & \hspace{1cm} \textstyle  u^i_k \; = \; {\text{arg\,min}}_{u^i\in \gamma_i^k \mathbb{U}_i} \; \|u^i\|,                                                                                      \\
        \begin{split}
            &\textstyle \frac{\partial}{\partial x^i}b_{\phi}^{ir}(t_k)(f^i_c + g^i_c u_k^i)  \geq  -\frac{\partial}{\partial t}b_{\phi}^{ir}(t_k) - \lambda b_{\phi}^{ir}(t_k)  \\ &\hspace{4.3cm} \textstyle - \nu^{ir}(t_k)
            -{}^{r}\epsilon_{\phi}^{ir}(t_k),
        \end{split}\label{eq: leader constraint} \\
        \begin{split}
            &\textstyle \frac{\partial}{\partial x^i}b_{\phi}^{i}(t_k)(f^i_c + g^i_c u_k^i) \geq  -\frac{\partial}{\partial t}b_{\phi}^{i}(t_k) - \lambda b_{\phi}^{i}(t_k) - \nu^{i}(t_k),
        \end{split}\label{eq:independednt slack constraint}             
    \end{align}
\end{subequations}
where we have omitted the arguments $x_k^i,x_k^r$ for brevity. The agent $r \in \mathcal{N}_{\psi}(i)$ corresponds to a single neighbor of agent $i$ such that $i$ is the leader of the edge $(i,r)\in \mathcal{E}_{\psi}$, while the factor $\gamma_i^k \in (0,1]$ is a shrinking factor that scales the control input set from $\mathbb{U}^i$ to $\gamma^i_k\mathbb{U}^i:= \{\gamma_k^iu^i| u^i \in \mathbb{U}^i\}$ and is adaptively computed based on the risk of not fulfilling the collaborative task $\phi^{ir}$. On the other hand ${}^{r}\epsilon_{\phi}^{ir}(t_k)$ is defined as $ {}^{r}\epsilon_{\phi}^{ir}(t_k) = \min_{u^r_k \in \gamma_{k}^r \mathbb{U}^r} \;  \frac{\partial}{\partial x^r}b_{\phi}^{ir}(t_k)(f^r_c + g^r_c u_k^r) $, and is communicated to agent $i$ by agent $r$ itself at each time $t_k$, while $\lambda$ is a positive constant. It was proved in \cite{gregcdc} that the trajectories $\bm{x}$ obtained by the closed-loop dynamics of the system under the decentralized control law \eqref{eq:new decentralized optimization with tasks} are such that $\bm{x} \models \psi$. We implemented \eqref{eq:new decentralized optimization with tasks} as a controller for a team of 5 drones connected by a star graph with edges $\mathcal{E}_{\psi} = \{(1,2),(1,3),(1,4),(1,5)\}$ such that agents $2,3,4$ and $5$ are leaders of their respective edge shared with agent $1$. The agents are then tasked with visiting two target points of interest while maintaining a time-varying reference formation. Each agent stores the specification $\phi^{i} \land \bigwedge_{j\in \mathcal{N}_{\psi}(i)} \phi^{ij}$ in the knowledge database before starting the mission. We interfaced the \symaware toolbox with the Pybullet \cite{cite:pybullet} environment for this simulation (Fig. \ref{fig:drones}). Details about the tasks and a video of the simulations can be found at~\cite{symaware2024framework}.
\subsection{Use Case III: Risk-Aware Dynamic Task Allocation}

\begin{figure}[htbp]
    \includegraphics[width=0.75\columnwidth,trim={0cm 0.35cm 0 0.3cm},clip]{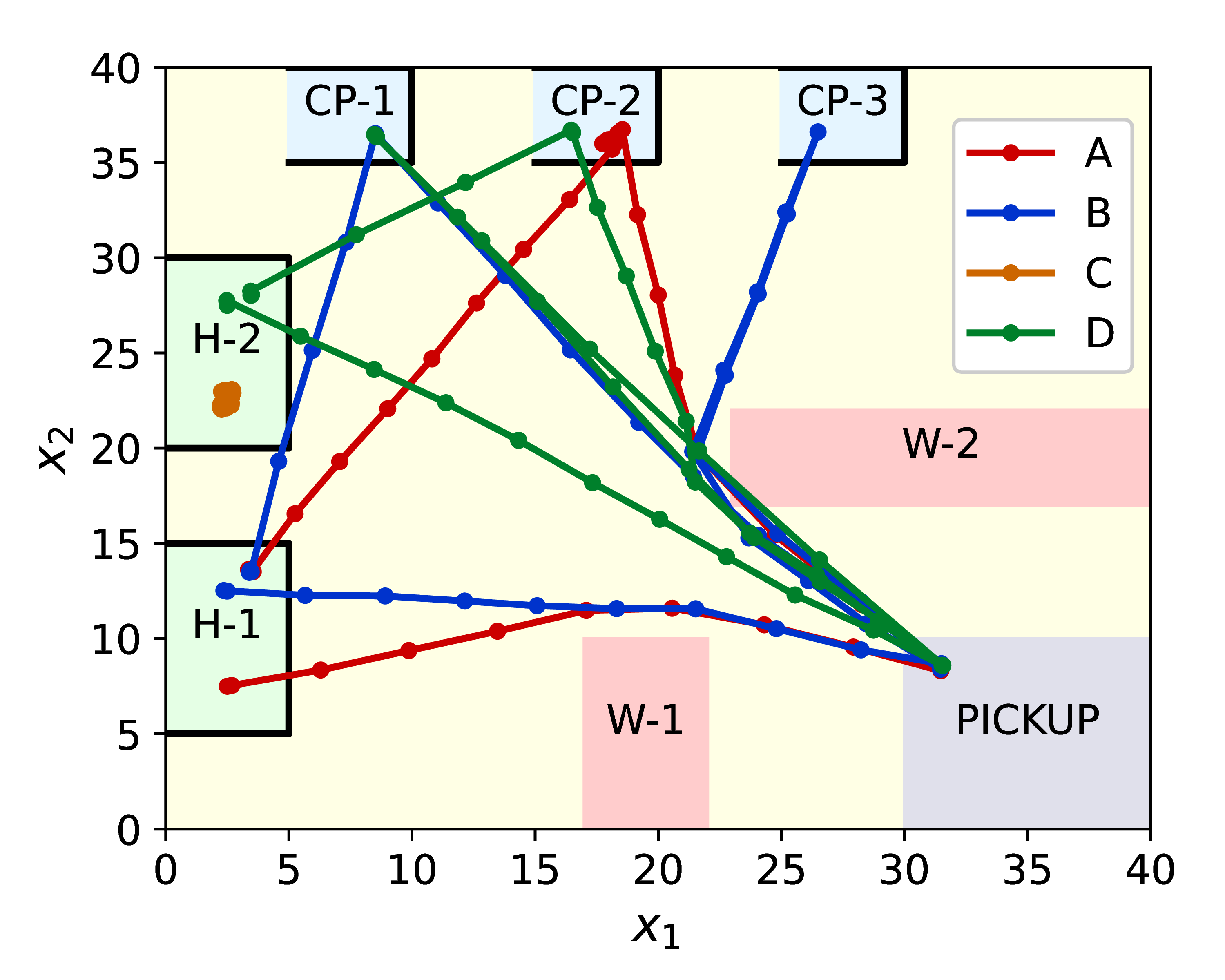}
    \centering
    \caption{An automated warehouse use case.}
    \label{Fig:warehouse}
\end{figure}

A \gls{mas} should automatically adjust its strategies according to the evolution of the environment based on the awareness of the potential risk. A typical example is an automated warehouse scenario as illustrated in Fig.~\ref{Fig:warehouse}, where four mobile robots are required to fetch packages from the collecting points (CP) to the pick-up point while avoiding collisions with the walls (W). The fetch tasks may be assigned during the runtime, meaning that each robot should evaluate its capability of taking the new fetch task given the accomplishment of its current job. Eventually, all robots should return home (H). Fig.~\ref{Fig:warehouse} shows an example where the tasks are described in the following natural language.
\textbf{1)} Robots A, B start from H-1 and C, D start from H-2. All robots should not leave the warehouse for all time.
\textbf{2)} At step $1$, a robot should fetch a package from CP-I to the pickup point within 10 steps. Meanwhile, another two robots should each fetch a package from CP-2 and CP-3, respectively, to the pickup point within 7 steps.
\textbf{3)} At step $15$, two robots should each fetch a package from CP-1 and CP-2, respectively, to PICKUP in 10 steps.
\textbf{4)} At step $30$, all robots should return home in 10 steps.\par

\symaware provides a systematic framework to calculate, communicate, and mitigate risk among agents. Fig.~\ref{Fig:warehouse} shows a solution ensuring all packages are successfully fetched despite the dynamic uncertainties of the robots. We can see that A fetches a package from CP-2, B fetches a package from CP-1 then another from CP-3, and D fetches packages from CP-2 and CP-1, respectively. C does not fetch any packages since it is configured to have the worst equipment, leading to a high risk of failing the task. This example showcases that \textit{SymAware} allows programming highly adaptive \gls{mas} that can accomplish complicated tasks in an uncertain environment. The mathematical theory supporting this result can be referred to in~\cite{engelaar2024risk}.

% \begin{remark}
%     In this experiment, the computation of control inputs by solving problem \eqref{Eq:TMPC} is distributed to each agent, leading to a computational complexity linear to the number of assigned specifications. In this sense, our proposed method allows for a comparable computational load of each agent to a conventional single-agent case, allowing it to be implemented in practical scenarios.
% \end{remark}

\section{CONCLUSIONS}

This paper introduces an applicable software framework to develop \gls{mas}. A rich set of specialized components allows each agent to conduct and communicate situational awareness efficiently. The modularized architecture enhances its extendability to generic applications. Future development plans include increasing the number of supported simulators and integrating components leveraging more aspects of \gls{sa} to achieve a more resilient agent. 

%\textcolor{red}{All the case studies have been added in the main} \href{https://gitlab.mpi-sws.org/sadegh/eicsymaware/-/tree/base?ref_type=heads#case-studies}{\textcolor{red}{README}} \textcolor{red}{and} \href{https://sadegh.pages.mpi-sws.org/eicsymaware/readme.html}{\textcolor{red}{documentation}} \textcolor{red}{so we can just link it once. Let me know if you want to change anything about how they are presented (link to repository, reference).}

% Zengjie: I have commented this command out to avoid the blanks in the reference

%%%%%%%%%%%%%%%%%%%%%%%%%%%%%%%%%%%%%%%%%%%%%%%%%%%%%%%%%%%%%%%%%%%%%%%%%%%%%%%%

%%%%%%%%%%%%%%%%%%%%%%%%%%%%%%%%%%%%%%%%%%%%%%%%%%%%%%%%%%%%%%%%%%%%%%%%%%%%%%%%

%%%%%%%%%%%%%%%%%%%%%%%%%%%%%%%%%%%%%%%%%%%%%%%%%%%%%%%%%%%%%%%%%%%%%%%%%%%%%%%%
%\section*{APPENDIX}


%\section*{ACKNOWLEDGMENT}

%%%%%%%%%%%%%%%%%%%%%%%%%%%%%%%%%%%%%%%%%%%%%%%%%%%%%%%%%%%%%%%%%%%%%%%%%%%%%%%%

\bibliography{root}

\end{document}

%% file: setup.tex
\IEEEoverridecommandlockouts % This command is only needed if you want to use the \thanks command
\usepackage{cite}
\usepackage[T1]{fontenc}
\usepackage[frozencache,cachedir=.]{minted}
\usepackage{amsmath,amssymb,amsfonts}
\usepackage{subcaption}
\usepackage{caption}
\usepackage{algorithmic}
\usepackage{graphicx}
\usepackage{textcomp}
\usepackage{xcolor}
\usepackage{xspace}
\usepackage{graphics} % for pdf, bitmapped graphics files
\usepackage{balance}
\usepackage{epsfig} % for postscript graphics files
\usepackage{mathptmx} % assumes new font selection scheme installed
\usepackage{times} % assumes new font selection scheme installed
\usepackage{tikz}
\usetikzlibrary{arrows.meta}

\usepackage{listings}
\usepackage[acronym]{glossaries}
\usepackage{hyperref} % Extensive support for hypertext in LATEX
\usepackage{adjustbox} % Adjusts the size of the table to fit the content
\usepackage{bm}
\graphicspath{ {fig/} }  % folder containing figures

\input{acronyms.tex}

\newcommand{\vect}[1]{\ensuremath{\boldsymbol{\mathrm{#1}}}}

\def\symaware{\textit{SymAware}\xspace}

%% file: acronyms.tex
\newglossaryentry{mas}
{
    name={Multi-Agent System},
    first={Multi-Agent System (MAS)},
    text={MAS},
    description={
            A multi-agent system is a system composed of multiple interacting agents.
            An agent is an entity capable of autonomous action in some environment in order to meet its design objectives.
        }
}
\newglossaryentry{aoc}{
    name={Agent-Oriented Computing},
    first={Agent-Oriented Computing (AOC)},
    text={AOC},
    description={
            Agent-Oriented Computing is a subfield of computer science that focuses on the design and implementation of systems composed of agents.
        }
}
\newglossaryentry{aose}{
    name={Agent-Oriented Software Engineering},
    first={Agent-Oriented Software Engineering (AOSE)},
    text={AOSE},
    description={
            Agent-Oriented Software Engineering is a subfield of software engineering that focuses on the design and implementation of software systems composed of agents.
        }
}
\newglossaryentry{oop}{
    name={Object-Oriented Programming},
    first={Object-Oriented Programming (OOP)},
    text={OOP},
    description={
            Object-Oriented Programming is a programming paradigm based on the concept of objects, which can hold a state in the form of fields, and methods, in the form of function with access to the object's state.
        }
}
\newglossaryentry{sa}{
    name={Situational Awareness},
    first={Situational Awareness (SA)},
    text={SA},
    description={
            Situational awareness is the main subject of this paper.
        }
}
\newglossaryentry{mpc}{
    name = {Model Predictive Control},
    first = {Model Predictive Control (MPC)},
    text = {MPC},
    description = {
            Model Predictive Control (MPC) is a powerful optimization-based control method that produces control policies by
            solving an optimal control problem at each discrete time instant based on the current system state over a finite,
            receding horizon.
        }
}
\newglossaryentry{adas}{
    name = {Advanced Driver Assistance Systems},
    first = {Advanced Driver Assistance Systems (ADAS)},
    text = {ADAS},
    description = {
            Advanced Driver Assistance Systems (ADAS) are systems developed to automate, adapt, and enhance vehicle systems for safety and better driving.
        }
}

\newglossaryentry{fts}{
    name = {Finite Transition System},
    first = {Finite Transition System (FTS)},
    text = {FTS},
    description = {
        A naive description of a Kripke structure
    }
}

\newglossaryentry{ltl}{
    name = {Linear-time Temporal Logic},
    first = {Linear-time Temporal Logic (LTL)},
    text = {LTL},
    description = {
        Linear-time Temporal Logic
    }
}

\newglossaryentry{stl}{
    name = {Signal Temporal Logic},
    first = {Signal Temporal Logic (STL)},
    text = {STL},
    description = {
        Signal Temporal Logic
    }
}

\newglossaryentry{api}{
    name = {Application Programming Interface},
    first = {Application Programming Interface (API)},
    text = {API},
    plural={APIs},
    firstplural={Application Programming Interfaces (APIs)},
    description = {
            An application programming interface (API) defines a way for two software components to communicate with each other.
        }
}

%% file: fig/main-loop.tex
% generated by Plantuml 1.2024.7       
\definecolor{plantucolor0000}{RGB}{255,255,255}
\definecolor{plantucolor0001}{RGB}{24,24,24}
\definecolor{plantucolor0002}{RGB}{0,0,0}
\definecolor{plantucolor0003}{RGB}{168,0,54}
\definecolor{plantucolor0004}{RGB}{226,226,240}
\definecolor{plantucolor0005}{RGB}{238,238,238}
\definecolor{plantucolor0006}{RGB}{254,255,221}
\begin{tikzpicture}[yscale=-1
,every node/.style={scale=2}
,pstyle0/.style={color=plantucolor0001,fill=white,line width=1.0pt}
,pstyle1/.style={color=black,line width=1.5pt}
,pstyle2/.style={color=plantucolor0001,line width=0.5pt,dash pattern=on 5.0pt off 5.0pt}
,pstyle3/.style={color=plantucolor0003,line width=1.0pt,dash pattern=on 1.0pt off 4.0pt}
,pstyle4/.style={color=plantucolor0001,fill=plantucolor0004,line width=0.5pt}
,pstyle6/.style={color=plantucolor0001,fill=plantucolor0001,line width=1.0pt}
,pstyle7/.style={color=plantucolor0001,line width=1.0pt}
,pstyle8/.style={color=plantucolor0001,fill=black,line width=1.5pt}
,pstyle9/.style={color=plantucolor0001,line width=1.0pt,dash pattern=on 2.0pt off 2.0pt}
,pstyle10/.style={color=plantucolor0001,fill=plantucolor0006,line width=0.5pt}
]
\draw[pstyle0] (262.372pt,265.3662pt) rectangle (272.372pt,446.1631pt);
\draw[pstyle0] (337.0071pt,323.6318pt) rectangle (347.0071pt,387.8975pt);
\draw[pstyle0] (608.4019pt,91.7021pt) rectangle (618.4019pt,208.2334pt);
\draw[pstyle1] (10pt,53.2969pt) rectangle (773.659pt,454.1631pt);
\draw[pstyle2] (164.919pt,36.2969pt) -- (164.919pt,216.2334pt);
\draw[pstyle3] (164.919pt,216.2334pt) -- (164.919pt,244.2334pt);
\draw[pstyle2] (164.919pt,244.2334pt) -- (164.919pt,471.1631pt);
\draw[pstyle2] (266.522pt,36.2969pt) -- (266.522pt,216.2334pt);
\draw[pstyle3] (266.522pt,216.2334pt) -- (266.522pt,244.2334pt);
\draw[pstyle2] (266.522pt,244.2334pt) -- (266.522pt,471.1631pt);
\draw[pstyle2] (341.222pt,36.2969pt) -- (341.222pt,216.2334pt);
\draw[pstyle3] (341.222pt,216.2334pt) -- (341.222pt,244.2334pt);
\draw[pstyle2] (341.222pt,244.2334pt) -- (341.222pt,471.1631pt);
\draw[pstyle2] (415.7921pt,36.2969pt) -- (415.7921pt,216.2334pt);
\draw[pstyle3] (415.7921pt,216.2334pt) -- (415.7921pt,244.2334pt);
\draw[pstyle2] (415.7921pt,244.2334pt) -- (415.7921pt,471.1631pt);
\draw[pstyle2] (613.1862pt,36.2969pt) -- (613.1862pt,216.2334pt);
\draw[pstyle3] (613.1862pt,216.2334pt) -- (613.1862pt,244.2334pt);
\draw[pstyle2] (613.1862pt,244.2334pt) -- (613.1862pt,471.1631pt);
\draw[pstyle4] (104.919pt,10pt) arc (180:270:5pt) -- (109.919pt,5pt) -- (220.522pt,5pt) arc (270:360:5pt) -- (225.522pt,10pt) -- (225.522pt,30.2969pt) arc (0:90:5pt) -- (220.522pt,35.2969pt) -- (109.919pt,35.2969pt) arc (90:180:5pt) -- (104.919pt,30.2969pt) -- cycle;
\node at (105.919pt,8pt)[below right,color=black]{Environment};
\draw[pstyle4] (235.522pt,10pt) arc (180:270:5pt) -- (240.522pt,5pt) -- (294.222pt,5pt) arc (270:360:5pt) -- (299.222pt,10pt) -- (299.222pt,30.2969pt) arc (0:90:5pt) -- (294.222pt,35.2969pt) -- (240.522pt,35.2969pt) arc (90:180:5pt) -- (235.522pt,30.2969pt) -- cycle;
\node at (236.522pt,8pt)[below right,color=black]{Entity};
\draw[pstyle4] (309.222pt,10pt) arc (180:270:5pt) -- (314.222pt,5pt) -- (369.7921pt,5pt) arc (270:360:5pt) -- (374.7921pt,10pt) -- (374.7921pt,30.2969pt) arc (0:90:5pt) -- (369.7921pt,35.2969pt) -- (314.222pt,35.2969pt) arc (90:180:5pt) -- (309.222pt,30.2969pt) -- cycle;
\node at (310.222pt,8pt)[below right,color=black]{Model};
\draw[pstyle4] (384.7921pt,10pt) arc (180:270:5pt) -- (389.7921pt,5pt) -- (442.4778pt,5pt) arc (270:360:5pt) -- (447.4778pt,10pt) -- (447.4778pt,30.2969pt) arc (0:90:5pt) -- (442.4778pt,35.2969pt) -- (389.7921pt,35.2969pt) arc (90:180:5pt) -- (384.7921pt,30.2969pt) -- cycle;
\node at (385.7921pt,8pt)[below right,color=black]{Agent};
\draw[pstyle4] (564.1862pt,10pt) arc (180:270:5pt) -- (569.1862pt,5pt) -- (657.6176pt,5pt) arc (270:360:5pt) -- (662.6176pt,10pt) -- (662.6176pt,30.2969pt) arc (0:90:5pt) -- (657.6176pt,35.2969pt) -- (569.1862pt,35.2969pt) arc (90:180:5pt) -- (564.1862pt,30.2969pt) -- cycle;
\node at (565.1862pt,8pt)[below right,color=black]{Controller};
\draw[pstyle0] (262.372pt,265.3662pt) rectangle (272.372pt,446.1631pt);
\draw[pstyle0] (337.0071pt,323.6318pt) rectangle (347.0071pt,387.8975pt);
\draw[pstyle0] (608.4019pt,91.7021pt) rectangle (618.4019pt,208.2334pt);
\draw[color=black,fill=plantucolor0005,line width=1.5pt] (10pt,53.2969pt) rectangle (85.1592pt,75.5693pt);
\draw[pstyle1] (10pt,53.2969pt) rectangle (773.659pt,454.1631pt);
\node at (25pt,48.2969pt)[below right,color=black]{\textbf{loop}};
\node at (106.1592pt,49.2969pt)[below right,color=black]{\textbf{[At each simulation step]}};
\draw[pstyle6] (596.4019pt,87.7021pt) -- (606.4019pt,91.7021pt) -- (596.4019pt,95.7021pt) -- (600.4019pt,91.7021pt) -- cycle;
\draw[pstyle7] (416.135pt,91.7021pt) -- (602.4019pt,91.7021pt);
\node at (423.135pt,64.5693pt)[below right,color=black]{compute\_and\_update()};
\draw[pstyle8] (776.159pt,120.085pt) ellipse (4pt and 4pt);
\draw[pstyle6] (759.159pt,116.835pt) -- (769.159pt,120.835pt) -- (759.159pt,124.835pt) -- (763.159pt,120.835pt) -- cycle;
\draw[pstyle9] (618.4019pt,120.835pt) -- (765.159pt,120.835pt);
\node at (625.4019pt,93.7021pt)[below right,color=black]{event "iterating"};
\draw[pstyle6] (353.0071pt,145.9678pt) -- (343.0071pt,149.9678pt) -- (353.0071pt,153.9678pt) -- (349.0071pt,149.9678pt) -- cycle;
\draw[pstyle7] (347.0071pt,149.9678pt) -- (607.4019pt,149.9678pt);
\node at (359.0071pt,122.835pt)[below right,color=black]{set\_control\_input(Result)};
\draw[pstyle8] (770.5837pt,178.3506pt) ellipse (4pt and 4pt);
\draw[pstyle6] (753.5837pt,175.1006pt) -- (763.5837pt,179.1006pt) -- (753.5837pt,183.1006pt) -- (757.5837pt,179.1006pt) -- cycle;
\draw[pstyle9] (618.4019pt,179.1006pt) -- (759.5837pt,179.1006pt);
\node at (625.4019pt,151.9678pt)[below right,color=black]{event "iterated"};
\draw[pstyle6] (427.135pt,204.2334pt) -- (417.135pt,208.2334pt) -- (427.135pt,212.2334pt) -- (423.135pt,208.2334pt) -- cycle;
\draw[pstyle9] (421.135pt,208.2334pt) -- (612.4019pt,208.2334pt);
\node at (433.135pt,181.1006pt)[below right,color=black]{OK};
\draw[pstyle6] (250.372pt,261.3662pt) -- (260.372pt,265.3662pt) -- (250.372pt,269.3662pt) -- (254.372pt,265.3662pt) -- cycle;
\draw[pstyle7] (165.2205pt,265.3662pt) -- (256.372pt,265.3662pt);
\node at (172.2205pt,238.2334pt)[below right,color=black]{step()};
\draw[pstyle8] (6.5pt,293.749pt) ellipse (4pt and 4pt);
\draw[pstyle6] (23.5pt,290.499pt) -- (13.5pt,294.499pt) -- (23.5pt,298.499pt) -- (19.5pt,294.499pt) -- cycle;
\draw[pstyle9] (17.5pt,294.499pt) -- (164.2205pt,294.499pt);
\node at (24pt,267.3662pt)[below right,color=black]{event "stepping"};
\draw[pstyle6] (325.0071pt,319.6318pt) -- (335.0071pt,323.6318pt) -- (325.0071pt,327.6318pt) -- (329.0071pt,323.6318pt) -- cycle;
\draw[pstyle7] (272.372pt,323.6318pt) -- (331.0071pt,323.6318pt);
\node at (279.372pt,296.499pt)[below right,color=black]{step()};
\draw[pstyle10] (160pt,331.6318pt) rectangle (562pt,361.6318pt);
\node at (166pt,331.6318pt)[below right,color=black]{Apply the last control input to the simulation};
\draw[pstyle6] (283.372pt,383.8975pt) -- (273.372pt,387.8975pt) -- (283.372pt,391.8975pt) -- (279.372pt,387.8975pt) -- cycle;
\draw[pstyle9] (277.372pt,387.8975pt) -- (341.0071pt,387.8975pt);
\node at (289.372pt,360.7646pt)[below right,color=black]{OK};
\draw[pstyle8] (12.0753pt,416.2803pt) ellipse (4pt and 4pt);
\draw[pstyle6] (29.0753pt,413.0303pt) -- (19.0753pt,417.0303pt) -- (29.0753pt,421.0303pt) -- (25.0753pt,417.0303pt) -- cycle;
\draw[pstyle9] (23.0753pt,417.0303pt) -- (164.2205pt,417.0303pt);
\node at (29.5753pt,389.8975pt)[below right,color=black]{event "stepped"};
\draw[pstyle6] (176.2205pt,442.1631pt) -- (166.2205pt,446.1631pt) -- (176.2205pt,450.1631pt) -- (172.2205pt,446.1631pt) -- cycle;
\draw[pstyle9] (170.2205pt,446.1631pt) -- (266.372pt,446.1631pt);
\node at (182.2205pt,419.0303pt)[below right,color=black]{OK};
\end{tikzpicture}

%% file: fig/asynchronous.tex
% generated by Plantuml 1.2024.3       
\definecolor{plantucolor0000}{RGB}{255,255,255}
\definecolor{plantucolor0001}{RGB}{24,24,24}
\definecolor{plantucolor0002}{RGB}{0,0,0}
\definecolor{plantucolor0003}{RGB}{226,226,240}
\definecolor{plantucolor0004}{RGB}{238,238,238}
\begin{tikzpicture}[yscale=-1
,every node/.style={scale=1.5}
,pstyle0/.style={color=plantucolor0001,fill=white,line width=1.0pt}
,pstyle1/.style={color=black,line width=1.5pt}
,pstyle2/.style={color=plantucolor0001,line width=0.5pt,dash pattern=on 5.0pt off 5.0pt}
,pstyle3/.style={color=plantucolor0001,fill=plantucolor0003,line width=0.5pt}
,pstyle5/.style={color=plantucolor0001,fill=plantucolor0001,line width=1.0pt}
,pstyle6/.style={color=plantucolor0001,line width=1.0pt}
,pstyle7/.style={color=plantucolor0001,line width=1.0pt,dash pattern=on 2.0pt off 2.0pt}
]
\draw[pstyle0] (353.9248pt,91.8418pt) rectangle (363.9248pt,205.6592pt);
\draw[pstyle0] (353.9248pt,234.9316pt) rectangle (363.9248pt,264.2041pt);
\draw[pstyle1] (10pt,53.2969pt) rectangle (400.7374pt,272.2041pt);
\draw[pstyle2] (50pt,36.2969pt) -- (50pt,289.2041pt);
\draw[pstyle2] (358.237pt,36.2969pt) -- (358.237pt,289.2041pt);
\draw[pstyle3] (20pt,10pt) arc (180:270:5pt) -- (25pt,5pt) -- (75.1143pt,5pt) arc (270:360:5pt) -- (80.1143pt,10pt) -- (80.1143pt,30.2969pt) arc (0:90:5pt) -- (75.1143pt,35.2969pt) -- (25pt,35.2969pt) arc (90:180:5pt) -- (20pt,30.2969pt) -- cycle;
\node at (27pt,12pt)[below right,color=black]{Agent};
\draw[pstyle3] (306.237pt,10pt) arc (180:270:5pt) -- (311.237pt,5pt) -- (406.6126pt,5pt) arc (270:360:5pt) -- (411.6126pt,10pt) -- (411.6126pt,30.2969pt) arc (0:90:5pt) -- (406.6126pt,35.2969pt) -- (311.237pt,35.2969pt) arc (90:180:5pt) -- (306.237pt,30.2969pt) -- cycle;
\node at (313.237pt,12pt)[below right,color=black]{Component};
\draw[pstyle0] (353.9248pt,91.8418pt) rectangle (363.9248pt,205.6592pt);
\draw[pstyle0] (353.9248pt,234.9316pt) rectangle (363.9248pt,264.2041pt);
\draw[color=black,fill=plantucolor0004,line width=1.5pt] (10pt,53.2969pt) -- (91pt,53.2969pt) -- (91pt,60.5693pt) -- (81pt,70.5693pt) -- (10pt,70.5693pt) -- (10pt,53.2969pt);
\draw[pstyle1] (10pt,53.2969pt) rectangle (400.7374pt,272.2041pt);
\node at (25pt,52pt)[below right,color=black]{\textbf{loop}};
\node at (106pt,54pt)[below right,color=black]{\textbf{[Asyncronously, for each component]}};
\draw[pstyle5] (341.9248pt,87.8418pt) -- (351.9248pt,91.8418pt) -- (341.9248pt,95.8418pt) -- (345.9248pt,91.8418pt) -- cycle;
\draw[pstyle6] (50.0571pt,91.8418pt) -- (347.9248pt,91.8418pt);
\node at (57.0571pt,74.5693pt)[below right,color=black]{\textbf{1}};
\node at (70.9683pt,74.6392pt)[below right,color=black]{await async\_compute\_and\_update()};
\draw[pstyle6] (321.9248pt,121.1143pt) -- (363.9248pt,121.1143pt);
\draw[pstyle6] (321.9248pt,121.1143pt) -- (321.9248pt,134.1143pt);
\draw[pstyle6] (321.9248pt,134.1143pt) -- (362.9248pt,134.1143pt);
\draw[pstyle5] (354.9248pt,130.1143pt) -- (364.9248pt,134.1143pt) -- (354.9248pt,138.1143pt) -- (358.9248pt,134.1143pt) -- cycle;
\node at (113.1122pt,103.8418pt)[below right,color=black]{\textbf{2}};
\node at (127.0233pt,103.9116pt)[below right,color=black]{await async\_compute -\textgreater  Result};
\draw[pstyle6] (321.9248pt,163.3867pt) -- (363.9248pt,163.3867pt);
\draw[pstyle6] (321.9248pt,163.3867pt) -- (321.9248pt,176.3867pt);
\draw[pstyle6] (321.9248pt,176.3867pt) -- (362.9248pt,176.3867pt);
\draw[pstyle5] (354.9248pt,172.3867pt) -- (364.9248pt,176.3867pt) -- (354.9248pt,180.3867pt) -- (358.9248pt,176.3867pt) -- cycle;
\node at (141.0939pt,146.1143pt)[below right,color=black]{\textbf{3}};
\node at (155.005pt,146.1841pt)[below right,color=black]{await async\_update(Result)};
\draw[pstyle5] (61.0571pt,201.6592pt) -- (51.0571pt,205.6592pt) -- (61.0571pt,209.6592pt) -- (57.0571pt,205.6592pt) -- cycle;
\draw[pstyle7] (55.0571pt,205.6592pt) -- (357.9248pt,205.6592pt);
\node at (67.0571pt,188.3867pt)[below right,color=black]{\textbf{4}};
\node at (80.9683pt,188.4565pt)[below right,color=black]{: OK};
\draw[pstyle5] (341.9248pt,230.9316pt) -- (351.9248pt,234.9316pt) -- (341.9248pt,238.9316pt) -- (345.9248pt,234.9316pt) -- cycle;
\draw[pstyle6] (50.0571pt,234.9316pt) -- (347.9248pt,234.9316pt);
\node at (57.0571pt,217.6592pt)[below right,color=black]{\textbf{5}};
\node at (70.9683pt,217.729pt)[below right,color=black]{await next\_loop()};
\draw[pstyle5] (61.0571pt,260.2041pt) -- (51.0571pt,264.2041pt) -- (61.0571pt,268.2041pt) -- (57.0571pt,264.2041pt) -- cycle;
\draw[pstyle7] (55.0571pt,264.2041pt) -- (357.9248pt,264.2041pt);
\node at (67.0571pt,246.9316pt)[below right,color=black]{\textbf{6}};
\node at (80.9683pt,247.0015pt)[below right,color=black]{OK};
\end{tikzpicture}